\documentclass[10pt,twocolumn,letterpaper]{article}

\usepackage{cvpr}              % To produce the CAMERA-READY version

\usepackage{graphicx}
\usepackage{amsmath}
\usepackage{amssymb}
\usepackage{booktabs}

\usepackage[pagebackref,breaklinks,colorlinks]{hyperref}
\newcommand\Tau{\mathcal{T}}

\usepackage[capitalize]{cleveref}
\crefname{section}{Sec.}{Secs.}
\Crefname{section}{Section}{Sections}
\Crefname{table}{Table}{Tables}
\crefname{table}{Tab.}{Tabs.}

\def\cvprPaperID{34} % *** Enter the CVPR Paper ID here
\def\confName{CVPR}
\def\confYear{2022}

\begin{document}

%%%%%%%%% TITLE - PLEASE UPDATE
\title{Out-Of-Distribution Detection In Unsupervised Continual Learning}

% \name{Jiangpeng He and Fengqing Zhu}
% \address{
% Elmore Family School of Electrical and Computer Engineering \\ 
% Purdue University, West Lafayette, Indiana, 47906, USA
% }

\author{Jiangpeng He\\
{\tt\small he416@purdue.edu}
\and
Fengqing Zhu\\
{\tt\small zhu0@purdue.edu}
\and
{Elmore Family School of Electrical and Computer Engineering, Purdue University, USA}
}

\maketitle

%%%%%% ABSTRACT
\begin{abstract}
 Unsupervised continual learning aims to learn new tasks incrementally without requiring human annotations. However, most existing methods, especially those targeted on image classification, only work in a simplified scenario by assuming all new data belong to new tasks, which is not realistic if the class labels are not provided. Therefore, to perform unsupervised continual learning in real life applications, an out-of-distribution detector is required at beginning to identify whether each new data corresponds to a new task or already learned tasks, which still remains under-explored yet. In this work, we formulate the problem for Out-of-distribution Detection in Unsupervised Continual Learning (OOD-UCL) with the corresponding evaluation protocol. In addition, we propose a novel OOD detection method by correcting the output bias at first and then enhancing the output confidence for in-distribution data based on task discriminativeness, which can be applied directly without modifying the learning procedures and objectives of continual learning. Our method is evaluated on CIFAR-100 dataset by following the proposed evaluation protocol and we show improved performance compared with existing OOD detection methods under the unsupervised continual learning scenario. 
 
 %  and reveal two major issues of naively performing OOD detection by directly using model's output including (i) the biased output value towards new tasks and (ii) the decrease of output confidence under continual learning scenario. Then we introduce a novel method to address both issues by correcting the output bias and enhancing confidence for in-distribution data. 
%  and outperforms existing OOD detection methods with large margin, performance for OOD detection compared with existing methods under the OOD-UCL scenario. 
\end{abstract}

%%%%%%%%% BODY TEXT
\section{Introduction}
\label{sec:intro}

Unsupervised continual learning is an emerging future learning system, capable of learning new tasks incrementally from unlabeled data. It requires neither static datasets nor human annotations compared with supervised offline learning. Existing methods study this problem under the assumption that all new data belongs to new tasks. We argue that if human annotation is not available as common in unsupervised scenario, we cannot know whether the unlabeled new data belongs to new or learned tasks. For example, an image-based mobile food recognition system should be able to distinguish new and learned food images first instead of blindly treating all of them as new food classes to perform unsupervised continual learning for update. Therefore, in order to make unsupervised continual learning work in practical problems, an out-of-distribution (OOD) detector should be required at the beginning of each incremental learning step to identify whether each data belongs to new or already learned tasks. However, the problem of OOD detection in continual learning still remains under-explored, \textit{i.e.} none of the existing OOD detection methods target for continual learning. 

% \begin{figure}[t]
% \begin{center}
%   \includegraphics[width=.8\linewidth]{intro_ood.png}
% %   \vspace{-0.7cm}
%   \caption{An example of out-of-distribution detection where \textbf{ID} refers to in-distribution data and \textbf{OOD} refers to out-of-distribution data.}
%   \label{fig:intro_ood}
% \end{center}
% \end{figure} 

The goal of OOD detection for image classification is to detect novel classes data. However, it becomes more challenging under continual learning scenario due to (1) the training data of learned tasks becomes unavailable; (2) we also need to address catastrophic forgetting problem~\cite{CF}. Most existing methods cannot be applied here because they either require all training data for already learned tasks to train an OOD detector~\cite{Mahalanobis,Gram_matrix_OOD,ODD_distance}, or they need to modify the training procedure and objectives~\cite{modify_training_1,modify_training_2,modify_training_3}, which may sacrifice the classification accuracy. Therefore, we focus on ``post-hoc" methods~\cite{OOD_survey} that can be directly applied on any trained classification models to perform OOD detection based on the output confidence, which has been widely adopted in real-world environments to avoid the need to access training data. 

The central idea of ``post-hoc" methods to perform OOD detection is to assign in-distribution (ID) data with higher confidence value $\textit{Conf}_{in}$ than the OOD data $\textit{Conf}_{out}$ based on the output vector where the confidence $\textit{Conf}$ is defined as the maximum of softmax output~\cite{MSP,ODIN} or the energy score~\cite{Energy_score}. The detection performance greatly depends on the difference value of output confidence between ID and OOD data $\mathcal{D}_c = \textit{Conf}_{in} - \textit{Conf}_{out}$ where higher $\mathcal{D}_c$ indicates better discrimination. However, there exists two major issues in continual learning scenario that can lead to the decrease of $\mathcal{D}_c$ including (1) the biased output value towards new classes as revealed in~\cite{BIC,mainatining}; (2) the decrease of output confidence compared with offline learning due to the objective of improving generalization ability to mitigate catastrophic forgetting~\cite{LWF,yuan2020revisitingkd}. Both issues can result in performance degradation for existing ``post-hoc" methods.

% Nevertheless, the performance of existing "post-hoc" methods struggles under unsupervised continual learning scenario due to the decrease of output confidence between in-distribution and out-of-distribution data, which will be analyzed in detail in Section. 

In this work, we first formulate the OOD detection in unsupervised continual learning scenario denoted as OOD-UCL and introduce the corresponding evaluation protocol. Then, we propose a novel OOD detection method that can address both issues mentioned above to achieve improved performance in unsupervised continual learning scenario. The main contributions are summarized as following. 
\begin{itemize}
    \item To the best of our knowledge, we are the first to formulate the problem and evaluation protocol for out-of-distribution detection in unsupervised continual learning (OOD-UCL), which remains under-explored.
    \item A novel method is introduced for OOD detection by correcting output bias and enhancing output confidence difference based on task discriminativeness.
    \item We conduct extensive experiments on the CIFAR-100~\cite{CIFAR} dataset to show the effectiveness of each component of our proposed method compared to existing works under OOD-UCL scenario.
    % and our results significantly outperforms existing work 
\end{itemize}

% 1. Describe the background of unsupervised continual learning 
% 2. Illustrate why we need out-of-distribution detector to perform unsupervised continual learning. 
% 3. Simply introduce the existing out-of-distribution methods and illustrate why we focus on "post-doc" methods. 
% 4. Use a figure to illustrate our proposed scenario and the difference with previous ideal setting. 
% 5. Show the main contribution of this paper

\section{Related Work}
\label{sec:related work}
We focus on image classification problem and we review the existing methods that are related to our work including (1) unsupervised continual learning; (2) OOD detection. 

\subsection{Unsupervised Continual Learning}
\label{subsec: rel:UCL}
Compared with supervised case, unsupervised continual learning has not received much attention~\cite{survey_2020}. Stojanov \textit{et al.}~\cite{stojanov2019incremental_CRIB} introduced an unsupervised object learning environment to learn a sequence of single-class exposures. In addition, CURL~\cite{CURL} and STAM~\cite{smith2019STAM} are proposed for task-free unsupervised continual learning where task boundary is not given. Based on existing supervised protocol~\cite{ICARL}, the most recent work~\cite{he2021unsupervised} proposed to use pseudo labels obtained based on cluster assignments to perform continual learning and show promising results on several benchmark datasets in unsupervised scenario. However, they only assume a simplified scenario where all the new data belong to new classes, which rarely happens in real life applications when the class labels are not available. Therefore, an OOD detector that can work under unsupervised continual learning scenario becomes indispensable.

\subsection{Out-of-distribution Detection}
\label{subsec: rel:OOD}
% Studies on image classification based out-of-distribution detection can be generally summarized into two categories that require or not to modify training procedure of original classification. 
As illustrated in Section~\ref{sec:intro}, we focus on image classification based OOD detection and analyze this problem in continual learning scenario where the training objective is more challenging. Therefore, we target on methods that can be applied to any trained classification model without modifying the training procedure, which is called ``post-hoc" methods~\cite{OOD_survey}. Existing ``post-hoc" methods are originated from~\cite{MSP}, which directly uses the maximum softmax probability as the confidence score to discriminate ID and OOD data. Then ODIN~\cite{ODIN} applies temperature scaling and input perturbation to amplify the confidence difference $D_c$ between ID and OOD data where a large temperature transforms the softmax score back to the logit space. Built on these insights, recent work~\cite{Energy_score} proposed to use energy score as output confidence for OOD detection, which maps the output to a scalar through a convenient log-sum-exp operator. However, none of the existing ``post-hoc" methods consider the two issues in continual learning scenario as illustrated in Section~\ref{sec:intro}, resulting in performance degradation.

\begin{figure}[t]
\begin{center}
  \includegraphics[width=1.\linewidth]{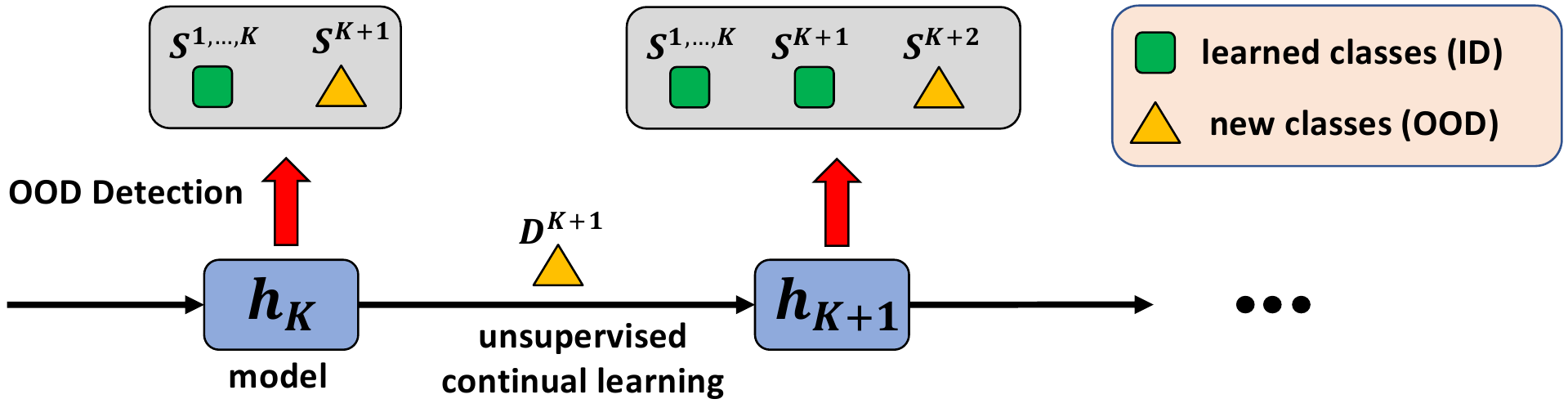}
  \vspace{-0.6cm}
  \caption{Formulation of out-of-distribution detection in unsupervised continual learning (OOD-UCL). $h_K$ refers to the updated incremental models after learning $\Tau^K$. $D^K$ and $S^K$ denote the corresponding training and testing splits for task $K$, respectively. }
  \label{fig:protocol}
    \vspace{-0.8cm}
\end{center}
\end{figure}

% \begin{figure*}[t]
% \begin{center}
%   \includegraphics[width=1.\linewidth]{comparison.png}
% %   \vspace{-0.7cm}
%   \caption{Implementing existing OOD detection methods under formulated OOD-UCL scenario using CIFAR-100 dataset with step size 10. The numerator and denominator of x-axis denote the number of learned classes (ID) and new classes (OOD). The upper-bound result is obtained in offline and supervised sceanrio. }
%   \label{fig:protocol}
% \end{center}
% \end{figure*} 

\section{Problem Formulation}
\label{sec:OOD-UCL}
% The objective is to discriminate between unlabeled learned tasks data and new task data, which can then be incorporated into any existing unsupervised continual learning methods to apply in real life applications. 
% The objective is to discriminate between unlabeled learned tasks data and new task data in continual learning scenario
The objective is to perform OOD detection in continual learning scenario to discriminate unlabeled learned tasks data (as ID) and new task data (as OOD), which can be then incorporated into any existing unsupervised continual learning methods to apply in real life applications. We formulate the out-of-distribution in unsupervised continual learning (OOD-UCL) problem based on the existing unsupervised class-incremental learning protocol~\cite{he2021unsupervised} to evaluate the OOD detection performance before each incremental learning step. Specifically, the continual learning for image classification problem $\Tau$ can be expressed as learning a sequence of $N$ tasks $\{\Tau^1,...,\Tau^N\}$ corresponding to $(N-1)$ incremental learning steps where the learning of the first task $\Tau^1$ is not included. Each task contains $M$ non-overlapped classes, which is known as incremental step size. Let $\{D^1, ..., D^N\}$ denote the training data and $\{S^1, ..., S^N\}$ denote the testing data for each task, we formulate the OOD-UCL with the following properties. 

\textbf{Property 1:} The OOD detection is performed at beginning of the learning step for each new task $\Tau^K$ where $K\in\{2,...N\}$. The test data belonging to learned tasks $S^i, i\in \{1,...K-1\}$ is regarded as ID data and the test data belonging to the current incremental step $S^K$ is regarded as the OOD data. Figure~\ref{fig:protocol} illustrates the evaluation protocol, where we perform total $(N-1)$ times OOD detection for continually learning a sequence of $N$ tasks $\{\Tau^1,...,\Tau^N\}$.

\textbf{Property 2:} The training data allowed for OOD detection before learning $\Tau^K$ is restricted to (1) the training set of $D^{K-1}$ and (2) the stored exemplars belonging to $\{\Tau^1,...,\Tau^{K-2}\}$ if applicable. This restricts the usage of most existing methods~\cite{Mahalanobis,Gram_matrix_OOD,ODD_distance} which requires all training data for learned classes to train an OOD detector.
% \textbf{Property 3:}
% The first property defines the evaluation protocol to perform $N-1$ times OOD detection for continually learning a sequence of $N$ tasks $\{\Tau^1,...,\Tau^N\}$. The second property restricts the usage of most existing OOD detection methods~\cite{Mahalanobis,Gram_matrix_OOD,ODD_distance} which requires all training data for learned classes to train an OOD detector. 

\textbf{Evaluation metrics:} In OOD detection, each test data is assigned with a confidence score where samples below the pre-defined confidence threshold are considered as OOD data. By regarding the ID data as positive and OOD data as negative, we can obtain a series of true positives rate (TPR) and false positive rate (FPR) by varying the thresholds. One of the commonly used metrics for OOD detection is \textbf{FPR95}, which measures the FPR when the TPR is $0.95$ and lower value indicates better detection performance. Besides, we can also calculate the area under receiver operating characteristic curve (\textbf{AUROC}~\cite{AUROC}) based on FPR and TPR as well as the area under the precision-recall curve (\textbf{AUPR}~\cite{AUPR}).  For both AUROC and AUPR, a higher value indicates better detection performance. 

% before each incremental learning step where the classes belonging to all learned tasks is regarded as ID data and the classes in the next incremental step, \textit{i.e.} , is regarded as OOD data. 

% Show why and how we set up this problem, what is the corresponding evaluation protocol

\begin{figure}[t]
\begin{center}
  \includegraphics[width=1.\linewidth]{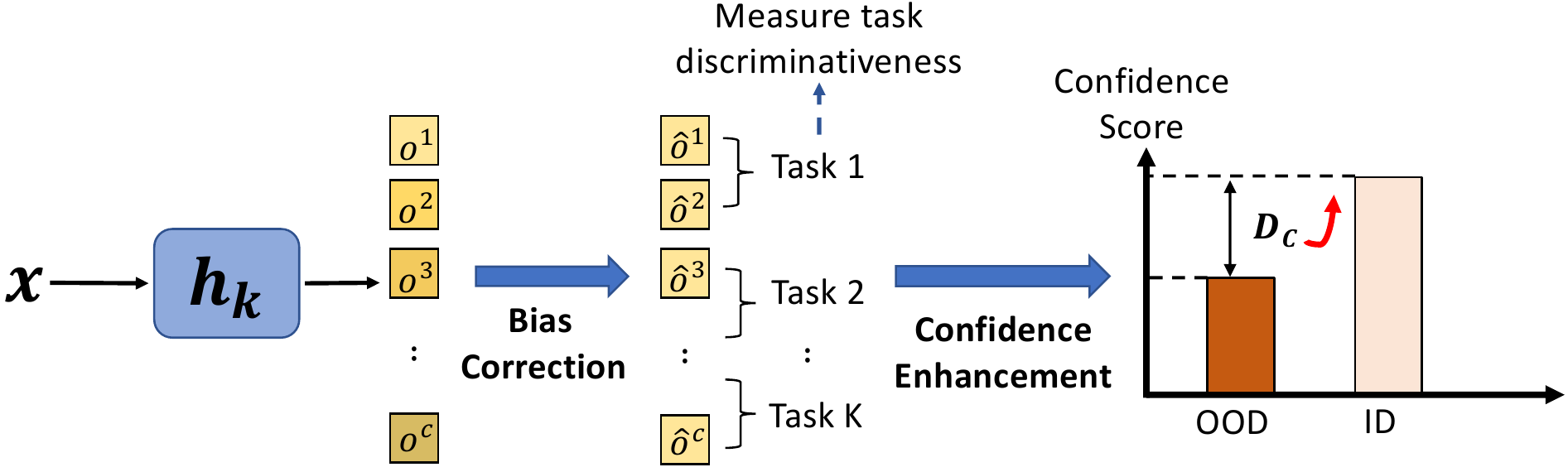}
  \vspace{-0.6cm}
  \caption{The overview of our proposed method where $\textbf{x}$ refers to input data and $h_K$ denotes the continual model after learning task $\Tau^K$. We first correct the bias of output $O$ to obtain $\hat{O}$ and then perform confidence enhancement to further increase the confidence difference $D_c$ to improve OOD detection performance. }
  \label{fig:our method}
    \vspace{-0.8cm}
\end{center}
\end{figure}

\begin{figure*}[t]
\begin{center}
  \includegraphics[width=.9\linewidth]{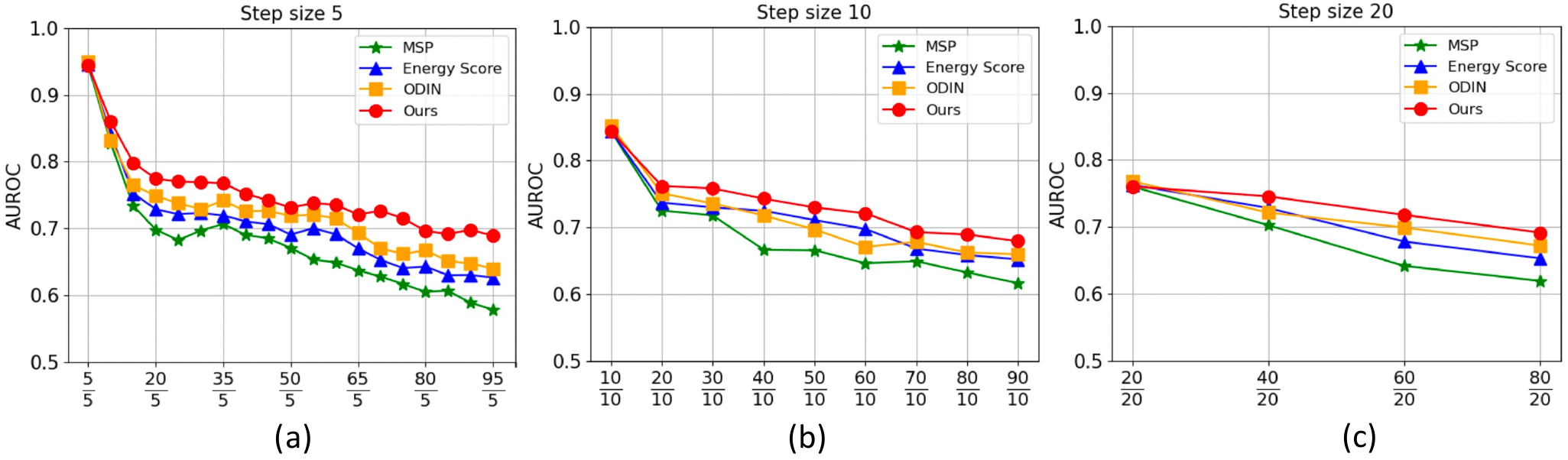}
  \vspace{-0.2cm}
  \caption{Results on CIFAR-100 with step size (a) 5 (b) 10 and (c) 20. The numerator and denominator of x-axis refers to the number of learned classes and new added classes, which are regarded as in-distribution and out-of-distribution data, respectively.}
  \label{fig:results}
  \vspace{-0.6cm}
\end{center}
\end{figure*} 

\begin{table*}[t!]
    \centering
    \scalebox{.92}
    {
    \begin{tabular}{|c|ccc|ccc|ccc|}
        \hline
        Methods & \multicolumn{3}{|c}{Step size 5} & \multicolumn{3}{|c|}{Step size 10} & \multicolumn{3}{c|}{Step size 20}  \\
        \hline
        & \textbf{AUROC}$\uparrow$ & \textbf{AUPR}$\uparrow$ & \textbf{FPR95}$\downarrow$ & \textbf{AUROC}$\uparrow$ & \textbf{AUPR}$\uparrow$ & $\textbf{FPR95}\downarrow$ & \textbf{AUROC}$\uparrow$ & \textbf{AUPR}$\uparrow$ & \textbf{FPR95}$\downarrow$ \\

        \hline
        MSP~\cite{MSP} & 0.679 & 0.947 & 0.855 & 0.685 & 0.899 & 0.873 & 0.681 & 0.834 & 0.877\\
        ODIN~\cite{ODIN} & 0.723 & 0.950 & 0.810 & 0.715 & 0.909 & 0.831 & 0.715 & 0.858 & 0.839\\
        Energy Score~\cite{Energy_score} & 0.707 & 0.950 & 0.824 & 0.714 & 0.907 & 0.837 & 0.706 & 0.853 & 0.844\\
        \hline
        \hline
        Ours (w/o BC) & 0.712 & 0.951 & 0.823 & 0.719 & 0.912 & 0.845 & 0.706 & 0.851 & 0.842\\
        Ours (w/o CE) & 0.708 & 0.947 & 0.836 & 0.713 & 0.907 & 0.851 & 0.699 & 0.844 & 0.854\\
        Ours & \textbf{0.754} & \textbf{0.959} & \textbf{0.793} & \textbf{0.736} & \textbf{0.915} & \textbf{0.824} & \textbf{0.729} & \textbf{0.874} & \textbf{0.814}\\
        \hline
    \end{tabular}
    }
    \caption{Average AUROC, AUPR and FPR95 on CIFAR-100 with step size 5, 10 and 20. BC and CE denotes bias correction step and confidence enhancement step, respectively. Best results are marked in bold.}
    \label{tab:cifar100}
    \vspace{-0.6cm}
\end{table*}
\vspace{-0.2cm}
\section{Our Method}
\vspace{-0.1cm}
\label{sec:our method}
In this section, we introduce a novel ``post-hoc" OOD detection method with the goal of improving the performance under unsupervised continual learning scenario, \textit{i.e.} increase the confidence difference $D_c$ between ID and OOD data for better discrimination. The overview of the proposed method is shown in Figure~\ref{fig:our method}, which can be directly applied without requiring any change to the existing classification models. There are two main steps including \textbf{bias correction} and \textbf{confidence enhancement} where we first correct the biased output value and then enhance the confidence difference $D_c$ based on task discriminativeness, which are described in Section~\ref{subsec: bias correction} and Section~\ref{subsec: confidence enhancement}, respectively.

\vspace{-0.1cm}
\subsection{Bias Correction}
\vspace{-0.1cm}
\label{subsec: bias correction}
Output bias towards new classes is a widely recognized issue~\cite{BIC,mainatining} caused by the lack of training data for learned tasks during continual learning. This results in the increase of the output value towards the biased classes for both ID and OOD data, therefore decreases the confidence difference $D_c$, \textit{i.e.} the degradation of OOD detection performance. Motivated by WA~\cite{mainatining} which shows the existence of biased weights in the FC classifier, we propose to perform bias correction by normalizing output logits based on the norm of weight vectors in the classifier corresponding to each learned class. Specifically, we denote the weight parameters in the classifier as $P \in \mathcal{R}^{d\times C}$ where $d$ is the dimension of extracted feature of each input sample and $C$ refers to the total number of classes seen so far. The weight norm of $P$ corresponds to each learned class is calculated as 
\begin{equation}
% \vspace{-0.2cm}
    |W^i| = L_2(P^{1,i},P^{2,i},...P^{d,i}), i\in\{1,2,...C\}
% \vspace{-0.2cm}
    \label{eq:weight-norm}
\end{equation}
where $L_2()$ denotes the $l_2$ normalization and $P^{j,k}$ refers to the element of $j^{th}$ row and $k^{th}$ column in $P$. Let $O = \{o^1, o^2, ..., o^C \}$ denote the output from the classifier, we normalize it through 
\begin{equation}
% \vspace{-0.2cm}
    \hat{o}^i = o^i/|W^i|, i\in\{1,2,...C\}
    % \vspace{-0.2cm}
    \label{eq:normalization}
\end{equation}
where $\hat{o}^i$ refers to the corrected output for class $i$. Our weight-based normalization generates the corrected output by efficiently mitigating the bias effect from the classifier. 

\vspace{-0.1cm}
\subsection{Confidence Enhancement}
\vspace{-0.1cm}
\label{subsec: confidence enhancement}
The learning objective also changes in continual learning scenario. Besides learning new tasks, we also need to maintain the learned knowledge. As shown in~\cite{pereyra2017pen_entropy}, higher confident output can decrease the model's generalization ability, which leads to catastrophic forgetting. 
% Since the learning objective in continual learning is to learn new tasks as well as maintaining the learned knowledge, higher output confidence can decrease the generalization ability of the network~\cite{pereyra2017pen_entropy}, which leads to catastrophic forgetting. 
Most existing continual learning methods address this problem by adding regularization to restrict the change of parameters~\cite{LWF,EEIL,ILIO,rebalancing,ICARL, He_2021_ICCVW} when learning new tasks, which decrease the output confidence for both ID and OOD data, resulting in the decrease of confidence difference $D_c$. Our goal is to increase $D_c$ to achieve better detection performance. Our proposed confidence enhancement method is motivated by the most recent work~\cite{He_2022_WACV, he2022_expfree}, which show that the continual learning model is able to maintain the discriminativeness within each learned task. Ideally, an ID data should be more confident and task-discriminative than OOD data. Therefore, after correcting the biased output, we apply softmax on $\hat{O} = \{\hat{o}^1, \hat{o}^2, ..., \hat{o}^C \}$ to obtain $\hat{S} = \{\hat{s}^1, \hat{s}^2, ..., \hat{s}^C \}$. We extract the maximum value as $\hat{S}_{max} = max(\hat{S})$ and its corresponding task index $I_{max} = \mathop{\textit{argmax}}_{i = 1,2...K} (\hat{S})$ where $K$ denotes the total number of tasks $\{\Tau^1,...,\Tau^K\}$ learned so far. The softmax output value for task $\Tau_{I_{max}}$ is extracted from $\hat{S}$ as $\hat{S}_{I_{max}} = \{\hat{s}^1_{I_{max}}, \hat{s}^2_{I_{max}}, ... \hat{s}^M_{I_{max}}\}$ where $M$ refers to the number of classes in each task, \textit{i.e.} the incremental step size. We then measure the discriminativeness based on entropy as in Equation~\ref{eq:entropy} where lower entropy $H$ indicates more discriminative. 

\begin{equation}
    H_{I_{max}} = \sum_{i=1}^M \hat{s}^i_{I_{max}} \times log_M(\hat{s}^i_{I_{max}})
    \label{eq:entropy}
\end{equation}

Finally, we calculate the confidence score as

\begin{equation}
    \textit{Conf} = \frac{\hat{S}_{max}}{H_{I_{max}}+\epsilon}
    \label{eq:ourconf}
\end{equation}
where $\epsilon = 0.00001$ is used for regularization. Test samples assigned with larger score is regarded as ID data.

\section{Experimental Results}
\label{sec:experimental results}
In this section, we show the effectiveness of our OOD detection method by applying the baseline in~\cite{he2021unsupervised} to perform unsupervised continual learning. The experimental results by incorporating other continual learning method~\cite{rebalancing, he2021unsupervised} are available in the supplementary Section~\ref{sec: addition exps}. We follow the proposed evaluation protocol by comparing the OOD detection results with existing ``post-hoc" methods including \textbf{MSP}~\cite{MSP}, \textbf{ODIN}~\cite{ODIN} and \textbf{Energy Score}~\cite{Energy_score}. We run each experiment 5 times and report the average results. The implementation detail of all existing methods can be found in the supplementary Section~\ref{sec: implementation detail}. 

We use the CIFAR-100~\cite{CIFAR} dataset and divide the 100 classes into splits of 20, 10 and 5 tasks with corresponding incremental step size 5, 10 and 20, respectively. Following the protocol in Section~\ref{sec:OOD-UCL}, we perform OOD detection at the beginning of each new task except the first one. 

\subsection{Results on CIFAR-100}
\label{subsec: results on CIFAR-100}
Table~\ref{tab:cifar100} shows the average OOD detection results on CIFAR-100 in terms of AUROC, AUPR and FPR95 as introduced in Section~\ref{sec:OOD-UCL}. We observe significant improvements for OOD detection in unsupervised continual learning scenario compared with existing ``post-hoc" methods. Besides, we also include \textbf{ours (w/o BC)} and \textbf{ours (w/o CE)} for ablation study where \textit{BC} and \textit{CE} denote bias correction and confidence enhancement steps as illustrated in Section~\ref{sec:our method}. Note that the MSP~\cite{MSP} can be regarded as \textbf{ours (w/o BC and CE)}. Thus, both BC and CE improves the detection performance compared with MSP and our method including both steps achieve the best performance. In addition, the AUROC on CIFAR-100 for each incremental step is shown in Figure~\ref{fig:results}. Our method outperforms existing approaches at each step especially with larger margins for smaller step size, as both output bias and confidence decrease problems become more severe due to the increasing number of incremental learning steps.

\section{Conclusion}
\label{sec:conclusion}
In this work, we first formulate the problem of out-of-distribution detection in unsupervised continual learning (OOD-UCL) and introduce the corresponding evaluation protocol. Then a novel OOD detection method is proposed by correcting output bias and enhancing confidence difference between ID and OOD data. Our experimental results on CIFAR-100 show promising improvements compared with existing methods for various step sizes. 

For future work, instead of splitting the dataset with non-overlapped classes, we will focus on unsupervised continual learning in a more realistic scenario where each new task may contain both new classes and learned classes data. Therefore, a more efficient method that can perform continual learning based on the output of OOD detection is needed for real life applications.

% Table~\ref{tab:bias-corre} shows the preliminary results regarding the average AUROC where we incorporate the proposed weight-based output normalization with existing "post-hoc" methods. 
% \begin{table}[t!]
%     \centering
%     \scalebox{.72}{
%     \begin{tabular}{|cccc|}
%         \hline
%         Methods & Step size 5 & Step size 10 & Step size 20  \\
%         \hline
%         \hline
%         % Accuracy(\%) & Avg & Last & Avg & Last & Avg & Last & Avg & Last  \\
%         % \hline
%         % \hline
%         MSP & 0.637 (0.713)& 0.639 (0.708) &0.634 (0.692)\\
%         MSP + bias correction & 0.665 & 0.671 & 0.678\\
%         \hline
%         ODIN & 0.644 (0.719)& 0.643 (0.711) &0.651 (0.698)\\
%         ODIN + bias correction & 0.673 & 0.675 & 0.681\\
%         \hline
%         Energy Score & 0.659 (0.724)& 0.651 (0.716) &0.657 (0.703)\\
%         Energy Score + bias correction & 0.680 & 0.677 & 0.684\\
        
%         \hline
%     \end{tabular}
%     }
%   % \vspace{-0.2cm}
%     \caption{Average AUROC on CIFAR-100 with step size 5, 10 and 20. The results in () denote the upper-bound.}
%     \label{tab:bias-corre}
% \end{table}
% Dataset: CIFAR-100 

% Evaluation metric: AUROC 

% OOD methods: MOP, ODIN, Energy score 

% Continual learning methods: LWF, EEIL, LUCIR, WA. 
% \section{Conclusion}
% \label{sec:conclusion}
% Review of our contribution and results...

% Future work includes conducting complete experiments by applying unsupervised continual learning methods using OOD detection results. 
%%%%%%%%% REFERENCES
{\small
\bibliographystyle{ieee_fullname}
\bibliography{egbib}
}

\strut
\newpage
\begin{table*}[t!]
    \centering
    \scalebox{.92}{
    \begin{tabular}{|c|ccc|ccc|ccc|}
        \hline
        Methods & \multicolumn{3}{|c}{Step size 5} & \multicolumn{3}{|c|}{Step size 10} & \multicolumn{3}{c|}{Step size 20}  \\
        \hline
        & \textbf{AUROC}$\uparrow$ & \textbf{AUPR}$\uparrow$ & \textbf{FPR95}$\downarrow$         & \textbf{AUROC}$\uparrow$ & \textbf{AUPR}$\uparrow$ & \textbf{FPR95}$\downarrow$         & \textbf{AUROC}$\uparrow$ & \textbf{AUPR}$\uparrow$ & \textbf{FPR95}$\downarrow$ \\
        \hline
        % Accuracy(\%) & Avg & Last & Avg & Last & Avg & Last & Avg & Last  \\
        % \hline
        % \hline
        baseline & 0.754 & 0.959 & 0.793 & 0.736 & 0.915 & 0.824 & 0.729 & 0.874 & 0.814\\
        LUCIR + Pseudo Labels & 0.762 & 0.963 & 0.786 & 0.742 & 0.921 & 0.819 & 0.734 & 0.879 & 0.806\\
        \hline
    \end{tabular}
    }
  % \vspace{-0.2cm}
    \caption{Average AUROC, AUPR and FPR95 on CIFAR-100 with step size 5, 10 and 20.}
    \label{tab:lucir}
\end{table*}

\strut
\newpage

\begin{figure}[t!]
\begin{center}
%   \vspace{-0.5cm}
  \includegraphics[width=1.\linewidth]{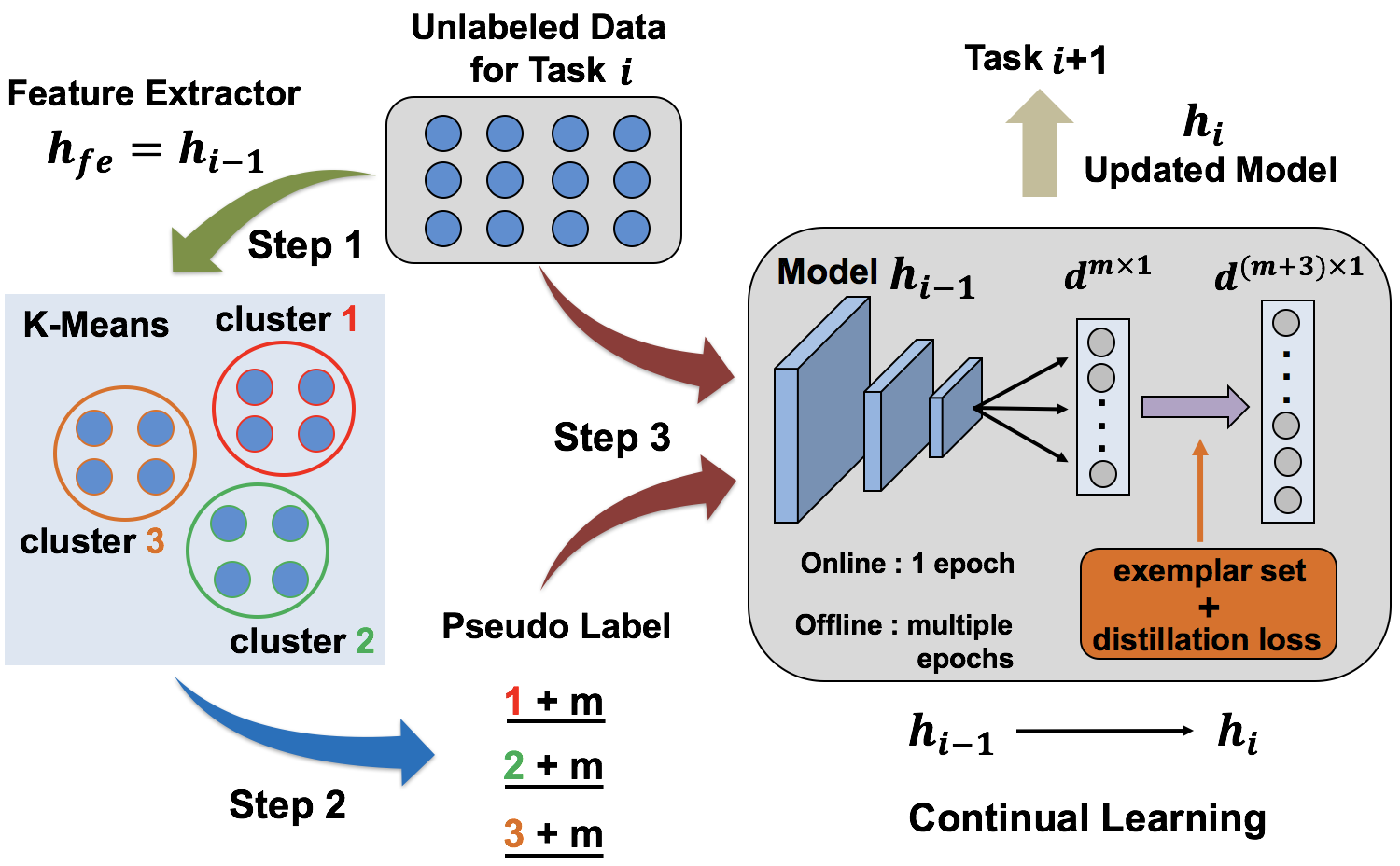}
  %\vspace{-0.2cm}
  \caption{\textbf{Overview of the baseline solution to learn the new task $i$.} \textbf{h} refers to the model in different steps and $\textbf{m}$ denotes the number of learned classes so far after task $i-1$. Firstly, the $\textbf{h}_{i-1}$ (except the last fully connected layer) is applied to extract feature embeddings used for K-means clustering where the number $\textbf{1}$, $\textbf{2}$, $\textbf{3}$ denote the corresponding cluster assignments. In step 2, the pseudo labels are obtained as $\textbf{1+m}$, $\textbf{2+m}$, $\textbf{3+m}$ respectively. Finally in step 3, the unlabeled data with pseudo label is used together for continual learning without requiring human annotations. }
%   \vspace{-.5cm}
  \label{fig:baselineUCL}
\end{center}
\end{figure}

\section{Implementation Detail}
\label{sec: implementation detail}
In this section, we provide the detail for methods implemented in experimental parts including (1) the method to perform unsupervised continual learning and (2) existing ``post-hoc" OOD detection methods, which will be illustrated in Section~\ref{subsec:UCL} and Section~\ref{subsec::OOD}, respectively. 
\subsection{Unsupervised Continual Learning}
\label{subsec:UCL}
In this work, we apply the baseline method proposed in~\cite{he2021unsupervised} to perform unsupervised continual learning as shown in Figure~\ref{fig:baselineUCL}, which includes three main steps: (1) Apply K-means clustering~\cite{K-MEANS} on extracted features for all new task data using lower layers of the continual learning model updated from last incremental step. (2) Obtain the pseudo labels based on the cluster assignments. (3) Perform unsupervised continual learning and maintain the learned knowledge by using exemplar set~\cite{ICARL} and knowledge distillation loss~\cite{KD}. 

In our experimental part, we follow the same setting to use ResNet-32~\cite{RESNET} for CIFAR-100~\cite{CIFAR} as the backbone network. The batch size of 128 with SGD optimizer, the initial learning rate is 0.1. We train 120 epochs for each incremental step and the learning rate is decreased by 1/10 for every 30 epochs. Exemplar size is set as $2,000$.

\subsection{OOD detection}
\label{subsec::OOD}
In our experiments, three existing "post-hoc" OOD detection methods are used for comparisons including MSP~\cite{MSP}, ODIN~\cite{ODIN} and Energy Score~\cite{Energy_score}. 

\textbf{MSP} directly applies the trained classification network to uses the maximum of softmax probability as the confidence score to discriminate between in-distribution and out-of-distribution data, which is regarded a strong baseline.
Specifically, for each input data $\textbf{x}$, we obtain the softmax output using trained classification model $\mathcal{F}_c$. The confidence score is calculated as $$\textit{Conf} = Max(Softmax(\mathcal{F}_c(\textbf{x})))$$ 

\textbf{ODIN} further improves the performance by introducing the temperature scaling and input data pre-processing. Specifically, they proposed to calculate softmax probability using output value scaled by temperature $T > 1$ and use the maximum as confidence score.

Besides, to increase the difference between in-distribution and out-of-distribution data, they pre-process each input data by adding small perturbation 
$$\Tilde{\textbf{x}} = \textbf{x} - \epsilon\textit{sign}(-\nabla_{\textbf{x}}log(\frac{\mathcal{F}_c(\textbf{x})}{T}))$$
Then, the final confidence score is given by $$\textit{Conf} = Max(Softmax(\frac{\mathcal{F}_c(\Tilde{\textbf{x}})}{T}))$$
In our experiment, we use $\epsilon = 0.001$ and $T=1,000$. 

\textbf{Energy Score} is proposed to replace the maximum of softmax probability as the confidence score. Specifically, considering the dimension of output logits $\mathcal{F}_c(\textbf{x})$ is $K$. The energy function is used to calculate the confidence score, which is defined as 
$$ \textit{Conf} = E(\textbf{x}) = -T\times log(\sum_i^Ke^{\mathcal{F}_c(\textbf{x})/T})$$
where we use $T=1,000$ as temperature scaling.

\section{Additional Experimental Results}
\label{sec: addition exps}
For all experiments shown in the paper, we apply the \textbf{baseline} in~\cite{he2021unsupervised} to perform unsupervised continual learning as illustrated in~\ref{subsec:UCL}. In this section, we will demonstrate that our proposed OOD detection method can work with other existing continual learning methods to achieve higher performance. Following~\cite{he2021unsupervised}, we apply \textbf{LUCIR~\cite{rebalancing} + Pseudo Labels} to perform unsupervised continual learning and the results are shown in Table~\ref{tab:lucir}.

%%%%%% ABSTRACT

% {\small
% \bibliographystyle{ieee_fullname}
% \bibliography{egbib}
% }

% \end{document}

\end{document}